\documentclass[11pt]{article}

\usepackage[final]{acl}
\usepackage{times}
\usepackage{latexsym}
\usepackage{amsmath}
\usepackage{amssymb}
\usepackage[T1]{fontenc}
\usepackage[utf8]{inputenc}
\usepackage{booktabs}
\usepackage{multirow}
\usepackage{microtype}
\usepackage{graphicx}
\usepackage{subcaption}
\usepackage{xspace}
\usepackage{pifont}
\usepackage{tcolorbox}
\usepackage{xcolor}
\usepackage{soul}
\usepackage{colortbl}
\usepackage{caption}
\usepackage{epstopdf}
\usepackage{tikz}
\usetikzlibrary{positioning, arrows.meta, shadows}
\usepackage{float}
\usepackage{enumitem}
\usepackage[linesnumbered,ruled,vlined]{algorithm2e}
\usepackage{arydshln}
\usepackage{inconsolata}
\usepackage{hyperref}

\definecolor{lightred}{RGB}{255, 245, 245}
\definecolor{lightblue}{RGB}{245, 250, 255}
\definecolor{darkblue}{RGB}{0, 0, 139}
\definecolor{taggray}{RGB}{100, 100, 100}
\definecolor{softred}{RGB}{190, 30, 30}
\definecolor{arrowcolor}{RGB}{80, 80, 80}
\definecolor{processblue}{RGB}{0, 50, 150}
\definecolor{customblue}{RGB}{65, 105, 225}
\definecolor{pastelblue1}{RGB}{230, 243, 255}
\definecolor{pastelblue2}{RGB}{189, 224, 254}
\definecolor{pastelblue3}{RGB}{147, 197, 253}
\definecolor{checkblue}{RGB}{59, 130, 246}

\newcommand{\cmark}{\ding{51}}
\newcommand{\xmark}{\ding{55}}

\newcommand{\scizoom}{\textsc{SciZoom}\xspace}
\newcommand{\prellm}{\textit{Pre-LLM}\xspace}
\newcommand{\postllm}{\textit{Post-LLM}\xspace}

\hypersetup{
    colorlinks=true,
    linkcolor=darkblue,
    filecolor=magenta,
    urlcolor=processblue,
    citecolor=darkblue,
}

\title{SciZoom: A Large-scale Benchmark for Hierarchical Scientific Summarization across the LLM Era}

\author{
  \textbf{Han Jang\textsuperscript{1,2,$\dagger$}},
  \textbf{Junhyeok Lee\textsuperscript{1,3,$\dagger$}},
  \textbf{Kyu Sung Choi\textsuperscript{1,2,3,$*$}}
\\
  \small \textsuperscript{1}Seoul National University \quad
  \textsuperscript{2}Seoul National University Hospital  \quad
  \textsuperscript{3}Seoul National University College of Medicine \\
  \small \textsuperscript{$\dagger$}Equal contribution \quad
  \textsuperscript{$*$}Corresponding author \\
  \small \texttt{\{hanjang, jhlee0619, ent1127\}@snu.ac.kr} \\
}

\begin{document}
\maketitle

\begin{abstract}

The explosive growth of AI research has created unprecedented information overload, significantly increasing the demand for scientific summarization at multiple levels of granularity beyond traditional abstracts.
While large language models~(LLMs) are increasingly adopted for summarization, existing benchmarks remain limited in scale, target only a single granularity, and predate the LLM era, leaving them unable to capture the distributional shifts introduced by widespread LLM-assisted writing since November 2022.
To bridge these gaps, we introduce \scizoom, a benchmark comprising 44,946 papers from four top-tier ML venues (NeurIPS, ICLR, ICML, EMNLP) spanning 2020 to 2025.
\scizoom provides three hierarchical summarization targets~(Abstract, Contributions, and TL;DR) achieving compression ratios up to 600:1, and is explicitly stratified into \prellm and \postllm eras so that models can be evaluated under realistic distributional shift.
Beyond summarization evaluation, \scizoom's temporal design enables corpus-level mining of scientific writing evolution.
Our linguistic analysis reveals striking shifts in phrase patterns (up to 10$\times$ for formulaic expressions) and rhetorical style (23\% decline in hedging), suggesting that LLM-assisted writing produces more confident yet homogenized prose.
Our code and dataset are publicly available on \href{https://github.com/janghana/SciZoom}{GitHub} and \href{https://huggingface.co/datasets/hanjang/SciZoom}{Hugging Face}, respectively.

\end{abstract}

\section{Introduction}
\label{sec:intro}

\begin{figure}[t]
    \centering
    \includegraphics[width=\linewidth]{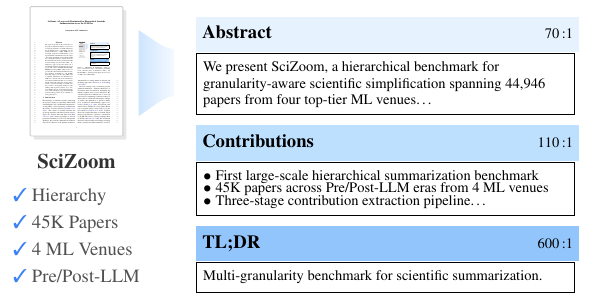}
    \caption{\textbf{Overview of \scizoom.} Three hierarchical granularity levels: Abstract~(70:1), Contributions~(110:1), and TL;DR~(600:1) across 44,946 papers from four top-tier ML venues.}
    \label{fig:overview}
\end{figure}

The landscape of scientific research is witnessing an explosive expansion, particularly within AI and Natural Language Processing~(NLP), a trend extensively documented in the science of science literature~\cite{fortunato2018science}.
The volume of scientific literature has grown exponentially, with indexed articles increasing by 47\% between 2016 and 2022 alone~\cite{hanson2024strain}, leaving researchers to face unprecedented information overload.
To mitigate this challenge, the research community has traditionally relied on abstracts as the primary medium for summarization, fostering numerous datasets and modeling approaches~\cite{cohan2018discourse, meng2021bringing}.

However, relying solely on abstracts presents significant limitations.
Abstracts themselves often remain dense and technical, demanding significant cognitive effort to parse effectively.
Researchers require more efficient ways to assess relevance before committing to full-text reading.
To address this, two additional elements have emerged as critical: the articulation of specific contributions, essential for understanding a paper's core novelty~\cite{teufel1999argumentative}, and extreme summaries such as TL;DR (Too Long; Didn't Read), now a required field in venues such as top-tier AI conferences~\cite{cachola2020tldr}.
Together, these form a natural hierarchy of scientific simplification, progressing from full text to abstract to contributions to TL;DR, each offering a progressively more compressed view of the paper's content.
The release of Large Language Models~(LLMs)~\cite{brown2020language} has accelerated this trend, with researchers increasingly leveraging them to generate and refine summaries at various granularities.

Yet LLMs are not limited to summarization.
Since the release of ChatGPT in November 2022, researchers have rapidly adopted these models for drafting, editing, and polishing full manuscripts themselves~\cite{liang2024monitoring}.
This widespread adoption means that any modern summarization benchmark must account for the resulting distributional shift, yet existing resources such as SciTLDR~\cite{cachola2020tldr} predate the LLM era entirely.
Beyond benchmarking, a temporally stratified corpus also enables an important line of inquiry: \textit{How have the linguistic patterns of scientific writing evolved across the LLM era?}
While \citet{liang2024monitoring} identified vocabulary shifts in peer reviews, no study has systematically mined how paper writing itself has changed across multiple granularities, precisely because no suitable resource existed.

To bridge these gaps, we introduce \textbf{\scizoom}, a large-scale benchmark designed primarily for hierarchical summarization under realistic distributional shift.
\scizoom curates 44,946 papers from four top-tier ML venues (NeurIPS~\cite{neurips}, ICLR~\cite{iclr}, ICML~\cite{icml}, EMNLP~\cite{emnlp}) spanning 2020 to 2025, explicitly partitioned into \prellm and \postllm eras around the November 2022 ChatGPT release.

The benchmark provides three hierarchical summarization targets at progressively higher compression ratios: Abstracts~(70:1), Contributions~(110:1), and TL;DRs~(600:1), with a three-stage extraction pipeline ensuring complete coverage across all granularity levels~(\S\ref{sec:benchmark}).
Leveraging \scizoom's temporal stratification, our linguistic analysis reveals up to 10$\times$ increases in formulaic phrases and a 23\% decline in hedging language in \postllm papers, providing the first large-scale empirical evidence that LLM-assisted writing produces more confident yet homogenized scientific prose.

Our contributions are as follows:

\begin{itemize}
    \item \scizoom is the first large-scale hierarchical summarization benchmark explicitly stratified by the LLM era, comprising 44,946 papers from four top-tier AI conferences spanning 2020--2025, with three target granularities (Abstract, Contributions, TL;DR) achieving compression ratios up to 600:1.
    \item A three-stage pipeline combining rule-based extraction, LLM validation, and generative synthesis achieves 100\% coverage for contribution extraction, while TL;DRs preserve author-written gold standards.
    \item Cross-era evaluation reveals significant distribution shifts between eras, and linguistic mining enabled by \scizoom's temporal design uncovers emergence of formulaic phrases~(up to 10$\times$) and 23\% reduction in hedging language, empirically characterizing generative AI's impact on ML/NLP scientific discourse.
\end{itemize}

\section{Related Work}
\label{sec:related}

\paragraph{Scientific Document Summarization.}
Early work on scientific summarization focused on abstract generation with discourse-aware attention~\cite{cohan2018discourse}, later extended by transformer-based~\cite{pilault2020extractive} and efficient attention approaches~\cite{huang2021efficient} for long documents.
The demand for rapid screening led to extreme summarization: SciTLDR~\cite{cachola2020tldr} introduced single-sentence summaries for 3,229 ICLR papers, and CiteSum~\cite{mao2022citesum} scaled to 93K papers using citation sentences.

More recent efforts target structured or long-form settings: SumSurvey~\cite{liu2024sumsurvey} provides long-document survey summarization across 18K surveys, and ACLSum~\cite{takeshita2024aclsum} offers aspect-based summaries for 250 NLP papers.
However, none jointly provide (i) hierarchical compression-based targets, (ii) large-scale single-paper coverage, and (iii) explicit Pre/Post-LLM temporal stratification.
\scizoom bridges this gap by providing structured annotations at multiple granularities (abstract, contributions, TL;DR), enabling fine-grained content mining within individual papers across temporal boundaries.

\paragraph{Knowledge Mining from Scientific Text.}
Argumentative Zoning~\cite{teufel1999argumentative} pioneered rhetorical role classification, and the SemEval-2021 NLPContributionGraph task~\cite{d2021semeval} formalized contribution extraction.
At the corpus level, S2ORC~\cite{lo2020s2orc} released 81M papers with full text, and OAG-Bench~\cite{zhang2024oag} introduced 10 academic graph mining tasks.
Most closely related, \citet{pramanick2025nature} analyze contribution types across 29K NLP papers, but focus on typological classification rather than summarization targets.
These efforts address static structures or inter-paper relationships without examining how writing evolves temporally.
\scizoom instead provides contributions as one of three hierarchical summarization granularities, paired with full text and temporal stratification, enabling generation evaluation and mining of writing evolution.

\paragraph{LLM Impact on Scientific Writing.}
Analysis of peer reviews from major ML conferences identified linguistic markers of LLM-assisted writing, with certain adjectives increasing 10--35$\times$ in frequency and an estimated 6.5\%--16.9\% of reviews containing substantial LLM-generated content~\cite{liang2024monitoring}.
A follow-up study found that approximately 22.5\% of computer science paper abstracts were substantially modified by LLMs by September 2024, compared with lower rates in mathematics~(7.7\%) and the Nature portfolio~(8.9\%)~\cite{liang2025quantifying}.

Despite these important findings, existing studies focus on vocabulary-level features and do not examine deeper patterns such as phrase structures or hedging behavior across different granularities.
\scizoom addresses this by providing hierarchical annotations across four granularity levels for 44,946 papers explicitly stratified into \prellm and \postllm eras.

\begin{table}[t]
\centering
\caption{Comparison with existing scientific summarization benchmarks.}
\label{tab:comparison}
\resizebox{\linewidth}{!}{%
\begin{tabular}{lccccccc}
\toprule
\multirow{2}{*}{Dataset} & \multirow{2}{*}{Papers} & \multirow{2}{*}{Years} & \multicolumn{4}{c}{Granularity} & Era \\
\cmidrule(lr){4-7}
 & & & Full & Abs & Contrib & TL;DR & Split \\
\midrule
SciTLDR~\cite{cachola2020tldr} & 3.2K & 2017--20 & \xmark & \cmark & \xmark & \cmark & \xmark \\
Multi-XScience~\cite{lu2020multi} & 40K & --2020 & \xmark & \cmark & \xmark & \xmark & \xmark \\
CiteSum~\cite{mao2022citesum} & 93K & --2020 & \xmark & \cmark & \xmark & \cmark & \xmark \\
\midrule
\textbf{\scizoom} & \textbf{45K} & \textbf{2020--25} & \cmark & \cmark & \cmark & \cmark & \cmark \\
\bottomrule
\end{tabular}
}%
\end{table}

\section{The \scizoom Benchmark}
\label{sec:benchmark}

This section describes the task formulation, data construction pipeline, and era-based stratification strategy.
Figure~\ref{fig:overview} provides an overview, and Table~\ref{tab:comparison} compares \scizoom with existing benchmarks.

\subsection{Task Formulation}

Scientific communication follows a natural simplification hierarchy.
We formalize this as four granularity levels $\mathcal{G} = \{G_1, G_2, G_3, G_4\}$: Full Text ($G_1$) contains complete paper content, Abstract ($G_2$) provides an author-written summary, Key Contributions ($G_3$) lists itemized core novelties, and TL;DR ($G_4$) offers an extreme summary for rapid browsing.

As shown in Table~\ref{tab:text_length}, this hierarchy spans from 12,263 words in full text to 20 words in TL;DR, achieving a compression ratio exceeding 600:1.
We define a hierarchical summarization task: given the full text ($G_1$), generate outputs at three target granularities, namely Abstract ($G_2$), Key Contributions ($G_3$), and TL;DR ($G_4$).
Cross-level consistency is evaluated post-hoc via entailment analysis (\S\ref{sec:experiments}).
For the TL;DR evaluation, we use the subset of 21,295 papers with author-provided ground truth, covering 47.4\% of the corpus.

\begin{figure}[t]
    \centering
    \includegraphics[width=\linewidth]{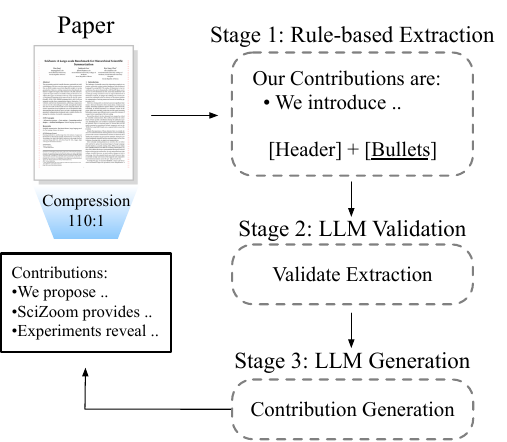}
    \caption{\textbf{Three-stage contribution extraction pipeline.} Rule-based extraction, LLM validation, and generative fallback achieve 100\% coverage.}
    \label{fig:construction}
\end{figure}

\begin{table}[t]
\centering
\caption{Average text length (words) by venue across granularity levels.}
\label{tab:text_length}
\resizebox{0.85\linewidth}{!}{%
\begin{tabular}{lrrrr}
\toprule
Venue & Full Paper & Abstract & Contrib & TL;DR \\
\midrule
NeurIPS~\cite{neurips} & 12,712 & 181 & 112 & 20 \\
ICLR~\cite{iclr} & 11,425 & 180 & 114 & 20 \\
ICML~\cite{icml} & 13,871 & 165 & 110 & 20 \\
EMNLP~\cite{emnlp} & 9,212 & 162 & 103 & 20 \\
\midrule
\textbf{Average} & \textbf{12,263} & \textbf{177} & \textbf{112} & \textbf{20} \\
\bottomrule
\end{tabular}
}%
\end{table}

\subsection{Data Construction Pipeline}

\begin{figure*}[t]
    \centering
    \includegraphics[width=\linewidth]{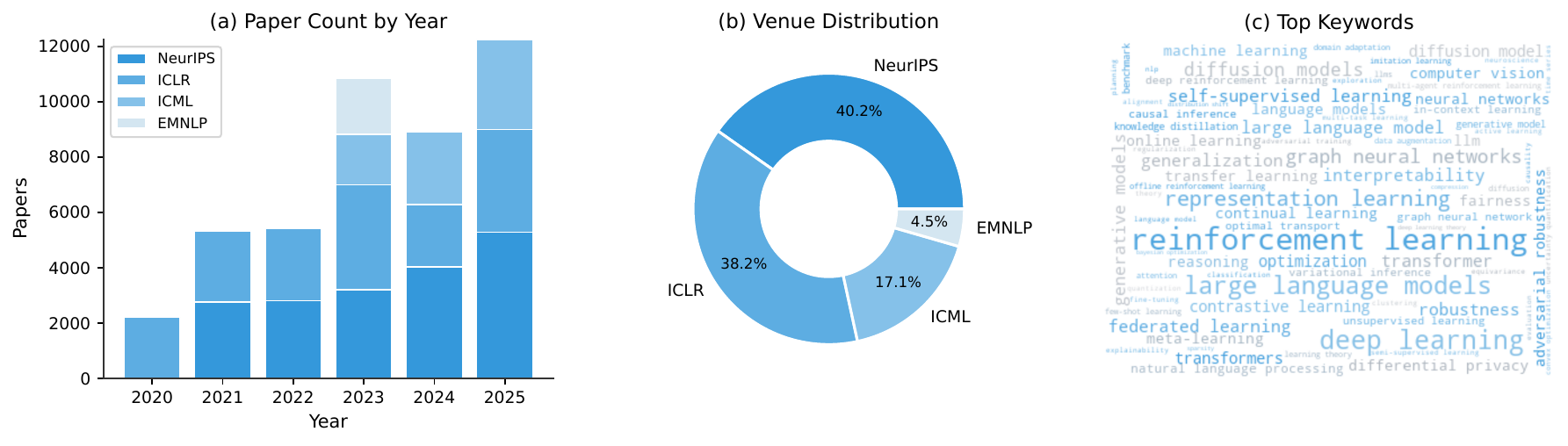}
    \caption{\textbf{Dataset distributions.} (a) papers by year across venues, (b) venue composition, (c) keyword word cloud.}
    \label{fig:dataset_dist}
\end{figure*}

We collected papers from four premier AI/ML venues spanning 2020 to 2025 via the OpenReview API.\footnote{\url{https://github.com/openreview/openreview-py}}
As detailed in Table~\ref{tab:dataset_overview}, our corpus comprises 44,946 papers.
Figure~\ref{fig:dataset_dist} shows the temporal growth trajectory across venues, with annual submissions increasing from 2,212 in 2020 to 12,246 in 2025; additional statistics are in Appendix~\ref{app:stats}.

\paragraph{Full Text and Abstract.}
We parsed PDF documents using PyMuPDF,\footnote{\url{https://pymupdf.readthedocs.io/}} extracting body text while removing abstracts, bibliographies, appendices, and formatting artifacts.
Abstracts are directly available as structured metadata in OpenReview submissions, ensuring 100\% coverage without additional processing.

\begin{table}[t]
\centering
\caption{Overview of the \scizoom dataset.}
\label{tab:dataset_overview}
\resizebox{0.7\linewidth}{!}{%
\begin{tabular}{lcrr}
\toprule
Venue & Years & Papers & TL;DR \\
\midrule
NeurIPS & 2021--2025 & 18,088 & 7,501 \\
ICLR & 2020--2025 & 17,154 & 10,492 \\
ICML & 2023--2025 & 7,695 & 2,200 \\
EMNLP & 2023 & 2,009 & 1,102 \\
\midrule
\textbf{Total} & 2020--2025 & \textbf{44,946} & \textbf{21,295} \\
\bottomrule
\end{tabular}
}%
\end{table}

\paragraph{Contribution Extraction Pipeline.}
Unlike abstracts and TL;DRs, key contributions are not provided as structured metadata.
We employ a three-stage cascading pipeline as illustrated in Figure~\ref{fig:construction}:
(1)~rule-based extraction identifies explicit contribution markers (e.g., ``Our contributions are:'') and extracts subsequent bulleted items;
(2)~LLM filtering uses Qwen3-4B-Instruct~\cite{yang2025qwen3} to verify whether the extracted items constitute valid contributions;
(3)~generative fallback synthesizes contributions from the abstract and introduction for papers where extraction fails.

The first two stages extracted author-written contributions from 15,397 papers (34\%), while the remaining 29,549 required LLM generation.
The quality of synthesized targets is validated through human evaluation in \S\ref{sec:experiments}.
We additionally release the author-written subset ($N{=}15{,}397$) as a clean evaluation slice (\S\ref{sec:experiments}); full pipeline details and prompt templates are in Appendix~\ref{app:pipeline}.

\paragraph{TL;DR Collection.}
TL;DRs are author-provided fields in OpenReview submissions.
As shown in Table~\ref{tab:dataset_overview}, 21,295 papers include TL;DRs, with higher availability in recent years due to venue policy changes.

\subsection{Dataset Analysis}
\label{sec:analysis}

\begin{figure}[t]
    \centering
    \begin{subfigure}[b]{0.48\linewidth}
        \includegraphics[width=\textwidth]{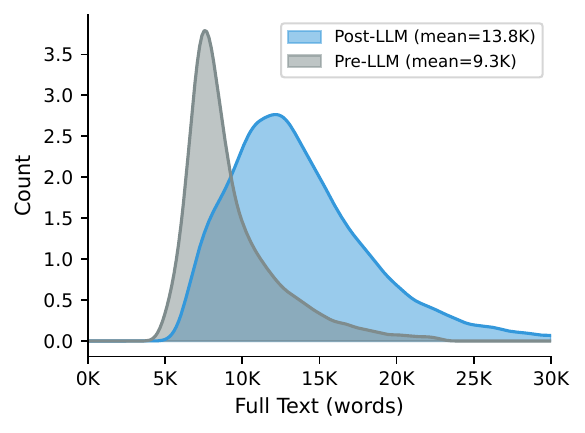}
        \caption{Full Text Length}
    \end{subfigure}
    \hfill
    \begin{subfigure}[b]{0.48\linewidth}
        \includegraphics[width=\textwidth]{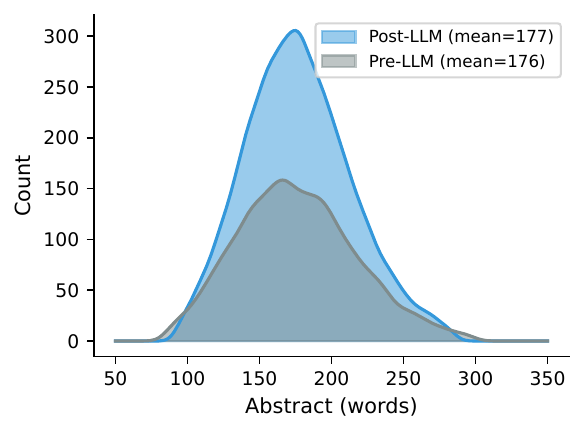}
        \caption{Abstract Length}
    \end{subfigure}
    \vspace{0.3cm}
    \begin{subfigure}[b]{0.48\linewidth}
        \includegraphics[width=\textwidth]{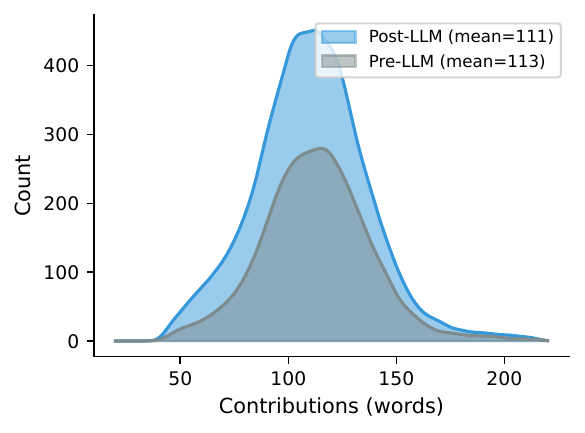}
        \caption{Contribution Length}
    \end{subfigure}
    \hfill
    \begin{subfigure}[b]{0.48\linewidth}
        \includegraphics[width=\textwidth]{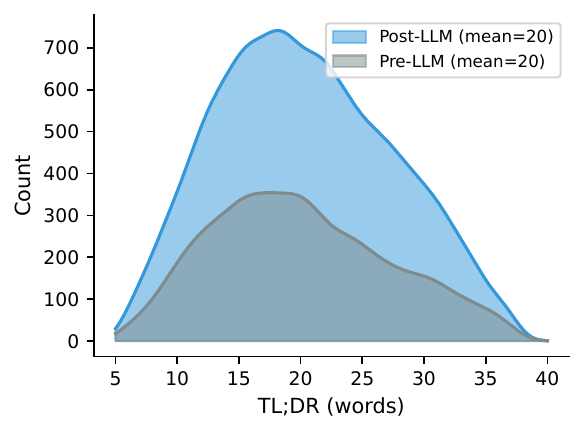}
        \caption{TL;DR Length}
    \end{subfigure}
    \caption{\textbf{Text length distributions.} Pre-LLM vs Post-LLM comparison across four granularity levels.}
    \label{fig:length_distribution}
\end{figure}

\paragraph{Pre/Post-LLM Era Stratification.}
A distinguishing feature of \scizoom is its explicit temporal stratification.
We define \prellm as papers submitted before ChatGPT's release on November 30, 2022; ICLR 2023 is classified as \prellm since its submission deadline was September 2022.
Table~\ref{tab:era_stats} compares paper characteristics across eras.
\postllm papers exhibit significantly longer full texts, averaging 13,940 words compared to 9,441 in \prellm, a 47.7\% increase (Figure~\ref{fig:length_distribution}a).
The TL;DR adoption rate also increased from 40.4\% to 51.6\% between eras.
Importantly, the number and length of claimed contributions remain stable across eras (4.3 vs.\ 4.2 items; 113 vs.\ 111 words), suggesting that structural norms around contribution claims have persisted even as writing style has shifted.

\begin{table}[t]
\centering
\caption{Comparison of paper characteristics between eras.}
\label{tab:era_stats}
\resizebox{0.9\linewidth}{!}{%
\begin{tabular}{lcc}
\toprule
Metric & \prellm & \postllm \\
\midrule
Papers & 16,754 (37.3\%) & 28,192 (62.7\%) \\
Full text (words) & 9,441 $\pm$ 3,770 & 13,940 $\pm$ 5,634 \\
Abstract (words) & 177 $\pm$ 44 & 177 $\pm$ 40 \\
Contributions (sentence) & 4.3 $\pm$ 0.8 & 4.2 $\pm$ 0.9 \\
Contributions (words) & 113 $\pm$ 30 & 111 $\pm$ 31 \\
TL;DR rate (\%) & 40.4 & 51.6 \\
\bottomrule
\end{tabular}
}%
\end{table}

\section{Experiments}
\label{sec:experiments}

\subsection{Experimental Setup}

We evaluate zero-shot LLMs on three summarization tasks: Abstract ($G_2$), Contribution ($G_3$), and TL;DR ($G_4$) generation from full paper text ($G_1$).

\paragraph{Evaluation Metrics.}
We adopt a complementary suite of metrics: ROUGE-L~\cite{lin2004rouge} and BLEU-4~\cite{papineni2002bleu} for lexical overlap, METEOR~\cite{banerjee2005meteor} for recall-oriented evaluation with synonym matching, and BERTScore~\cite{zhang2019bertscore} for semantic equivalence beyond surface forms.
Full results including ROUGE-1/2 are reported in Appendix~\ref{app:cross_era}.

\subsection{Baselines}

We benchmark six open-source LLMs in zero-shot settings, pairing older and newer versions within each family: Mistral-7B v0.1/v0.3~\cite{jiang2023mistral7b}, Llama-3/3.1-8B-Instruct~\cite{grattafiori2024llama}, and Qwen/Qwen2-7B~\cite{bai2023qwen,yang2024qwen2technicalreport}.
DeepSeek-R1~\cite{guo2025deepseek} (671B) serves as a frontier reference.
Note that \postllm models may have seen some test papers during pretraining, making our evaluation conservative.

All models generate outputs independently per granularity using greedy decoding (512 tokens for abstracts/contributions, 64 for TL;DRs) with front-back truncation to each model's context length; full details are in Appendix~\ref{app:preproc}.
For embedding-based analysis, we employ NV-Embed-v2~\cite{lee2025nv} for semantic similarity measurement and cross-granularity retrieval.

\begin{table*}[t]
\centering
\caption{Zero-shot performance on \scizoom benchmark (\%). Models are grouped into \postllm (newer) and \prellm (older) releases. DeepSeek-R1 (671B) is included as a frontier reference. \textbf{Bold}: best, \underline{underline}: second-best per column.}
\label{tab:main_results}
\resizebox{\textwidth}{!}{%
\begin{tabular}{ll|cccc|cccc|cccc}
\toprule
& & \multicolumn{4}{c|}{\textbf{Abstract}} & \multicolumn{4}{c|}{\textbf{Contribution}} & \multicolumn{4}{c}{\textbf{TL;DR}} \\
& \textbf{Model} & R-L & BL4 & MTR & BS & R-L & BL4 & MTR & BS & R-L & BL4 & MTR & BS \\
\midrule
\multirow{3}{*}{\rotatebox{90}{\small\postllm}}
& Mistral-7B-v0.3  & \underline{23.5} & \textbf{4.6} & \textbf{20.8} & 86.7 & 23.7 & 3.6 & 16.5 & 86.2 & 28.1 & \underline{4.9} & 27.3 & \underline{89.0} \\
& Llama-3.1-8B     & 22.4 & 3.4 & 18.7 & 86.8 & 25.7 & 4.6 & \underline{18.8} & \underline{86.9} & \underline{28.6} & \textbf{5.2} & \underline{27.8} & \textbf{89.1} \\
& Qwen2-7B         & 23.3 & 3.7 & 19.9 & \underline{87.0} & \textbf{29.8} & \textbf{6.6} & \textbf{22.6} & \textbf{87.6} & 26.3 & 4.0 & 23.1 & 88.4 \\
\midrule
\multirow{3}{*}{\rotatebox{90}{\small\prellm}}
& Mistral-7B-v0.1  & 18.1 & 2.8 & 14.7 & 85.2 & 23.8 & \underline{4.7} & 18.7 & 86.1 & 25.2 & 4.5 & \textbf{28.8} & 88.2 \\
& Llama-3-8B       & \textbf{23.8} & \underline{4.1} & \underline{20.3} & \textbf{87.1} & 24.4 & 3.7 & 16.2 & 86.4 & \textbf{28.8} & \textbf{5.2} & 27.5 & \underline{89.0} \\
& Qwen-7B          & 18.6 & 2.9 & 16.0 & 85.3 & 15.4 & 1.1 & 9.8 & 84.2 & 25.3 & 4.4 & 25.7 & 88.2 \\
\midrule
\multicolumn{1}{l}{Frontier}
& DeepSeek-R1      & 22.6 & 3.1 & 18.9 & 86.9 & \underline{25.6} & 2.9 & 18.2 & 87.2 & 24.0 & 3.1 & 20.4 & 88.0 \\
\bottomrule
\end{tabular}}
\end{table*}

\subsection{Main Results}

Table~\ref{tab:main_results} presents zero-shot performance across three summarization tasks, with models grouped by release era.
A qualitative comparison in Appendix Figure~\ref{fig:qualitative_comparison} confirms that Qwen2-7B most closely reproduces the ground-truth contribution structure.

\paragraph{Abstract Generation.}
\postllm models (Mistral-v0.3, Llama-3.1, Qwen2) and the \prellm Llama-3-8B achieve comparable performance with R-L scores around 23\% and BERTScore around 87\%.
However, older \prellm models show substantial degradation: Mistral-v0.1 drops to 18.1\% and Qwen-7B to 18.6\% R-L.
This 5+ percentage point gap suggests that model updates have significantly improved abstract generation capabilities.

\paragraph{Contribution Extraction.}
Qwen2-7B substantially outperforms all other models, achieving 29.8\% R-L compared to 25.7\% for Llama-3.1-8B and 23.7\% for Mistral-v0.3.
This suggests that Qwen2 better captures the bullet-point structure typical of contribution statements.
The performance gap is even more pronounced against \prellm models, with Qwen-7B achieving only 15.4\% R-L.
Interestingly, Mistral-v0.1 (23.8\%) performs comparably to its newer version v0.3 (23.7\%) on this task, suggesting that model updates do not uniformly improve all capabilities.

\paragraph{TL;DR Generation.}
Llama-3-8B and Llama-3.1-8B achieve the highest R-L scores with 28.8\% and 28.6\% respectively, followed closely by Mistral-v0.3 at 28.1\%.
The relatively lower performance across all models reflects the inherent challenge of extreme compression, where 12,000+ words must be distilled into approximately 20 words.
BERTScore remains high at 89\% across models, suggesting successful semantic preservation despite limited lexical overlap.

\paragraph{Frontier Reference (DeepSeek-R1).}
DeepSeek-R1 (671B) lies within the band of strong 7-8B baselines on R-L (22.6/25.6/24.0 for Abstract/Contrib/TL;DR) and BERTScore (86.9/87.2/88.0), and is not the best model on any task.
Its lower lexical overlap reflects a verbose reasoning style rather than a quality gap, as semantic-level scores (BERTScore, and the CIDEr/SciBERTScore/BARTScore in Appendix~\ref{app:domain_metrics}) remain comparable.
This indicates \scizoom is not saturated by simple scaling and remains challenging for frontier reasoning models.

\subsection{Benchmark Validation}

\paragraph{Human Evaluation of LLM-Generated Contributions.}
To assess the faithfulness of the 66\% LLM-generated contributions, two ML/NLP annotators independently rated 50 randomly sampled papers (223 individual items) against the source paper on a three-point scale: \textit{Supported}, \textit{Partially supported}, or \textit{Unsupported}.
Table~\ref{tab:human_eval} reports item and paper-level results.
No contribution was judged \textit{Unsupported} by either annotator (0\% hallucination rate), and 91.7\% were rated fully \textit{Supported}.
Inter-annotator agreement reaches $\kappa{=}0.50$ (moderate~\cite{mchugh2012interrater}), with all disagreements confined to the \textit{Supported}/\textit{Partially supported} boundary, reflecting numerical imprecisions rather than factual errors.

\begin{table}[t]
\centering
\caption{Human evaluation of LLM-generated contributions. Item-level (N=223) and paper-level (N=50, rated by worst item).}
\label{tab:human_eval}
\resizebox{\linewidth}{!}{%
\begin{tabular}{l|ccc|ccc}
\toprule
& \multicolumn{3}{c|}{\textbf{Item-level}} & \multicolumn{3}{c}{\textbf{Paper-level}} \\
\textbf{Rating} & A & B & Avg & A & B & Avg \\
\midrule
Supported          & 89.2 & 94.2 & 91.7 & 58.0 & 76.0 & 67.0 \\
Partially sup.     & 10.8 & \phantom{0}5.8 & \phantom{0}8.3 & 42.0 & 24.0 & 33.0 \\
Unsupported        & \phantom{0}0.0 & \phantom{0}0.0 & \phantom{0}0.0 & \phantom{0}0.0 & \phantom{0}0.0 & \phantom{0}0.0 \\
\midrule
Cohen's $\kappa$~\cite{cohen1960coefficient}   & \multicolumn{3}{c|}{0.50} & \multicolumn{3}{c}{--} \\
\bottomrule
\end{tabular}}
\end{table}

\paragraph{Author-written vs.\ LLM-Generated Contributions.}
To assess whether evaluating against the 66\% LLM-generated contribution targets introduces circularity, we split the corpus into the author-written subset ($N{=}15{,}397$) and the LLM-generated subset ($N{=}29{,}549$) and re-run all six baselines.
As shown in Table~\ref{tab:auth_vs_gen_main}, all six models score \emph{higher} on author-written targets ($\Delta$R-L $+0.02$ to $+0.19$), confirming that LLM-generated targets are not systematically easier.
Model rankings are fully preserved across both subsets, and the human evaluation confirms 91.7\% of LLM-generated items are fully supported with 0\% unsupported.
We release the author-written subset as a clean evaluation slice suitable for circularity-sensitive analyses.

\begin{table}[t]
\centering
\caption{Contribution-task performance on author-written vs.\ LLM-generated targets. Model rankings are preserved across both subsets.}
\label{tab:auth_vs_gen_main}
\resizebox{\linewidth}{!}{%
\begin{tabular}{l|cc|cc}
\toprule
& \multicolumn{2}{c|}{\textbf{ROUGE-L (\%)}} & \multicolumn{2}{c}{\textbf{BERTScore}} \\
\textbf{Model} & Auth & LLM & Auth & LLM \\
\midrule
Qwen2-7B & 42.1 & 23.3 & 0.890 & 0.869 \\
Llama-3.1-8B & 31.8 & 22.5 & 0.873 & 0.866 \\
Llama-3-8B & 33.3 & 19.7 & 0.873 & 0.858 \\
Mistral-7B-v0.3 & 30.2 & 20.4 & 0.867 & 0.859 \\
Mistral-7B-v0.1 & 29.0 & 21.1 & 0.866 & 0.859 \\
Qwen-7B & 16.6 & 14.7 & 0.842 & 0.842 \\
\bottomrule
\end{tabular}}
\end{table}

\paragraph{Cross-Level Entailment.}
To verify that model outputs remain logically consistent across granularity levels, we ran a DeBERTa-large-MNLI~\cite{he2020deberta,williams2018broad} entailment classifier over all model-generated (Abstract, Contributions, TL;DR) triples.
Table~\ref{tab:entailment_main} reports Abstract$\rightarrow$TL;DR entailment, the most aggressive compression pair.
Entailment exceeds $89\%$ across all five evaluated models, and contradiction stays at or below $0.4\%$; hierarchical outputs almost never introduce conflicting claims across compression levels.
Full Abstract$\rightarrow$Contribution entailment results are reported in Appendix~\ref{app:entailment}.

\begin{table}[t]
\centering
\caption{Cross-level entailment for Abstract$\rightarrow$TL;DR (\%). Contradiction $\leq$0.4\% across all models, including the frontier reference DeepSeek-R1.}
\label{tab:entailment_main}
\resizebox{\linewidth}{!}{%
\begin{tabular}{lccc}
\toprule
\textbf{Model} & \textbf{Entail} & \textbf{Contradict} & \textbf{Neutral} \\
\midrule
Mistral-7B-v0.3      & 95.4 & 0.1 & \phantom{0}4.5 \\
Qwen2-7B             & 93.6 & 0.1 & \phantom{0}6.2 \\
Llama-3-8B           & 93.6 & 0.1 & \phantom{0}6.3 \\
Llama-3.1-8B         & 90.3 & 0.1 & \phantom{0}9.5 \\
DeepSeek-R1          & 89.4 & 0.4 & 10.2 \\
\bottomrule
\end{tabular}}
\end{table}

\begin{table}[t]
\centering
\caption{Emerging formulaic trigrams in Post-LLM scientific abstracts. Document frequency indicates the number of papers containing each phrase. Pre/Post-LLM corpora contain 16,754 and 28,192 papers, respectively.}
\label{tab:trigrams}
\resizebox{0.95\linewidth}{!}{%
\begin{tabular}{lccr}
\toprule
\textbf{Trigram} & \textbf{Pre-LLM} & \textbf{Post-LLM} & \textbf{Ratio} \\
\midrule
\textit{novel framework that}         & 25 & 345 & 8.2$\times$ \\
\textit{recent advancements in}       & 18 & 288 & 9.5$\times$ \\
\textit{extensive experiments across}  & 17 & 273 & 9.5$\times$ \\
\textit{address this gap}             & 25 & 262 & 6.2$\times$ \\
\textit{building on this}             & 20 & 260 & 7.7$\times$ \\
\textit{emerged as promising}         & 27 & 202 & 4.4$\times$ \\
\textit{analysis reveals that}        & 30 & 194 & 3.8$\times$ \\
\textit{existing methods often}       & 10 & 174 & 10.3$\times$ \\
\bottomrule
\end{tabular}
}%
\end{table}

\subsection{Linguistic Analysis}
\label{sec:linguistic}

\scizoom's temporal stratification enables systematic mining of how scientific writing has evolved across the LLM era.

\paragraph{Emergence of Formulaic Phrases.}
Table~\ref{tab:trigrams} reveals significant increases in templatic expressions across \postllm abstracts.
The most striking increases appear in methodological framing: \textit{existing methods often} rose 10.3$\times$, while \textit{recent advancements in} and \textit{extensive experiments across} each increased 9.5$\times$, far exceeding the 1.68$\times$ baseline expected from corpus size difference alone.
A Mann-Whitney U test confirms the shift is statistically significant ($U{=}2.09{\times}10^{8}$, $p{<}10^{-177}$, $r{=}0.117$); per-venue tests on NeurIPS and ICLR are also significant at $p{<}10^{-15}$ (Appendix~\ref{app:per_venue}), ruling out venue composition as a confound.

\paragraph{Reduced Hedging Language.}
Following Hyland's lexicon of epistemic hedges in scientific writing~\cite{hyland1998hedging}, hedging language (\textit{may}, \textit{might}, \textit{could}, \textit{possibly}, \textit{potentially}, \textit{suggest}, \textit{indicate}, \textit{appear}, \textit{seem}, \textit{likely}) decreased by 22.8\%, dropping from 1.88 to 1.45 occurrences per 1,000 words.
In contrast, assertive language (\textit{demonstrate}, \textit{achieve}, \textit{outperform}, \textit{prove}, \textit{show}, \textit{establish}, \textit{confirm}) remained virtually unchanged ($-$0.4\%; 9.29 to 9.25 per 1,000 words).
A Mann-Whitney U test confirms the hedging shift is highly significant ($p{<}10^{-63}$), with consistent declines per-venue (NeurIPS: $-31.1\%$; ICLR: $-20.7\%$, both $p{<}10^{-15}$; Appendix~\ref{app:per_venue}).

This asymmetry is informative: if driven by evolving research norms, one would expect assertive language to increase in tandem.
Instead, selective removal of uncertainty markers is consistent with LLM-assisted writing stripping epistemic caution in ML/NLP abstracts, although we do not claim this characterizes all scientific prose; generalization to other disciplines remains future work.

\begin{figure}[t]
    \centering
    \includegraphics[width=\linewidth]{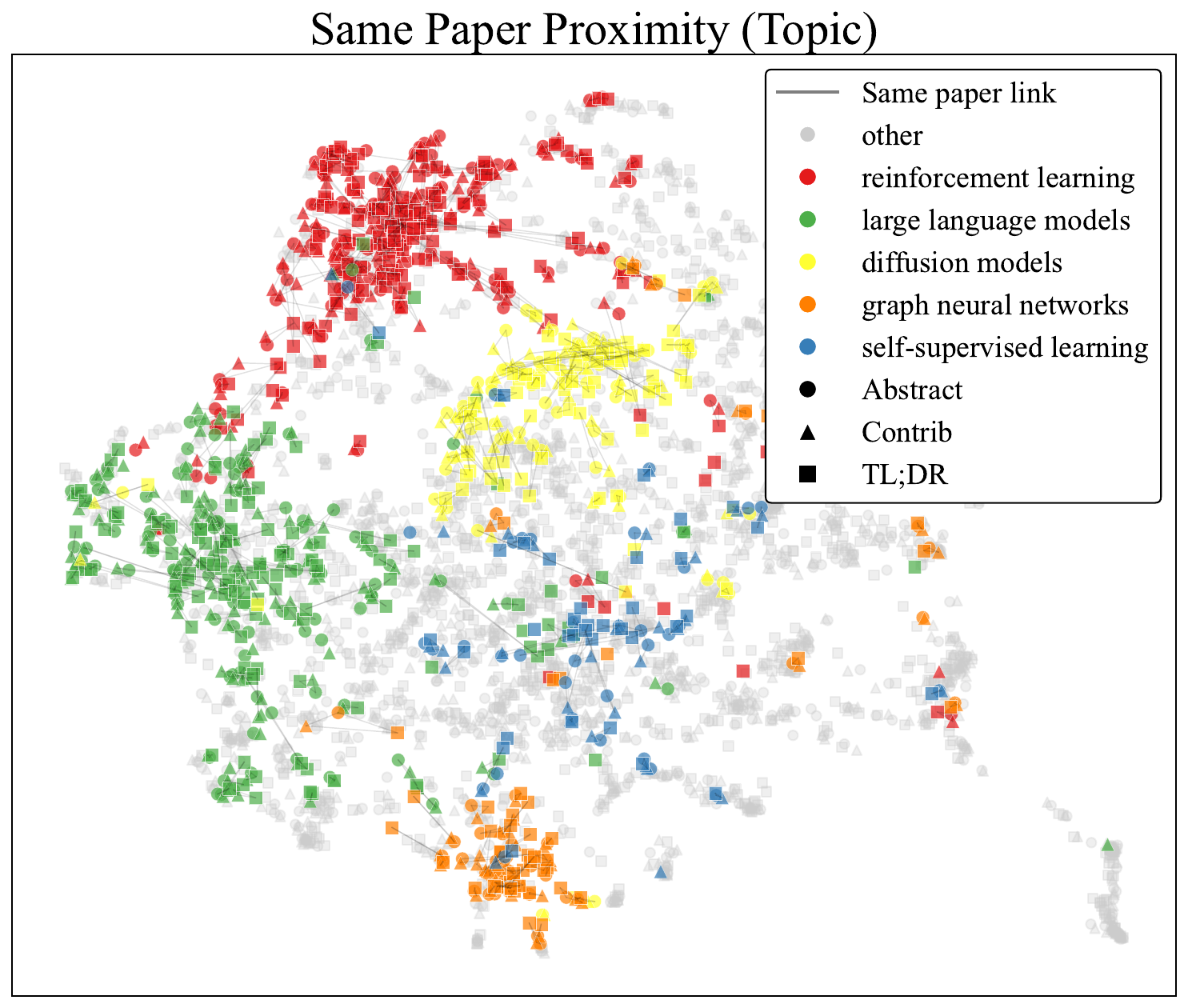}
    \caption{\textbf{UMAP visualization of hierarchical summaries.} Papers cluster by topic rather than granularity level, with an average intra-paper distance of 0.237.}
    \label{fig:umap_semantic}
\end{figure}

\begin{table}[t]
\centering
\caption{Ground-truth cross-granularity retrieval.}
\label{tab:retrieval_gt}
\resizebox{\linewidth}{!}{%
\begin{tabular}{l|cccc}
\toprule
\textbf{Direction} & \textbf{R@1} & \textbf{R@5} & \textbf{R@10} & \textbf{MRR} \\
\midrule
tldr$\rightarrow$abs    & 81.3 & 92.0 & 94.4 & 86.1 \\
tldr$\rightarrow$cont   & 72.3 & 85.3 & 88.8 & 78.3 \\
cont$\rightarrow$abs    & 98.9 & 99.6 & 99.8 & 99.2 \\
cont$\rightarrow$tldr   & 78.7 & 89.5 & 92.2 & 83.7 \\
abs$\rightarrow$tldr    & 79.1 & 90.5 & 93.1 & 84.3 \\
abs$\rightarrow$cont    & 95.0 & 97.7 & 98.3 & 96.2 \\
\bottomrule
\end{tabular}}
\end{table}

\subsection{Semantic Coherence}
\label{sec:coherence}

For a hierarchical benchmark, summaries at different granularities must preserve the core meaning of the same paper; we verify this via cross-granularity retrieval.
Table~\ref{tab:retrieval_gt} presents ground-truth cross-granularity retrieval: Contribution$\rightarrow$Abstract reaches 98.9\% R@1, and even TL;DR$\rightarrow$Abstract achieves 81.3\% R@1, demonstrating semantic preservation at 600$\times$ compression.
Model-generated summaries maintain similar coherence, with Qwen2-7B achieving 99.3\% R@1 for TL;DR$\rightarrow$Abstract (full results in Appendix~\ref{app:retrieval_def}).

Figure~\ref{fig:umap_semantic} further confirms this: papers cluster by topic rather than granularity level, with an average intra-paper distance of 0.237.
Table~\ref{tab:embedding_sim} shows that Pre/Post-LLM gaps in embedding similarity remain below 1\% across all models, confirming stable annotation quality across temporal boundaries.

\begin{table}[t]
\centering
\caption{Embedding similarity (\%) between generated and ground-truth summaries. Pre: \prellm subset, Post: \postllm subset.}
\label{tab:embedding_sim}
\resizebox{\linewidth}{!}{%
\begin{tabular}{l|cc|cc|cc}
\toprule
& \multicolumn{2}{c|}{\textbf{Abstract}} & \multicolumn{2}{c|}{\textbf{Contrib}} & \multicolumn{2}{c}{\textbf{TL;DR}} \\
\textbf{Model} & Pre & Post & Pre & Post & Pre & Post \\
\midrule
Mistral-7B-v0.1~\cite{jiang2023mistral7b} & 85.6 & 85.7 & 78.8 & 78.5 & \underline{77.8} & 78.2 \\
Llama-3-8B~\cite{grattafiori2024llama} & \textbf{86.3} & \underline{86.8} & \underline{79.1} & \underline{79.4} & \textbf{78.3} & \textbf{78.8} \\
Qwen2-7B ~\cite{yang2024qwen2technicalreport} & \textbf{86.3} & \textbf{87.2} & \textbf{83.2} & \textbf{84.0} & 77.6 & \underline{78.6} \\
\bottomrule
\end{tabular}}
\end{table}

\subsection{Era-based Analysis}
\label{sec:era_analysis}

We evaluated all models separately on \prellm and \postllm subsets.
\postllm models show stable or improved performance (Qwen2-7B gains +2.3 R-L on contributions), while older \prellm models degrade notably (Mistral-v0.1: $-$3.6, Qwen-7B: $-$4.1 on abstracts).
This asymmetry suggests that newer models have adapted to post-ChatGPT writing patterns, highlighting the importance of temporal stratification in benchmark design.

\section{Discussion}
\label{sec:discussion}

\paragraph{Implications for ML/NLP Scientific Communication.}
Our findings reveal measurable homogenization in ML/NLP abstracts following LLM adoption.
Within this scope, the 4--10$\times$ increase in formulaic phrases, combined with a 22.8\% decline in hedging language while assertive language remains stable, is consistent with LLM assistance systematically stripping epistemic caution rather than adding confidence, raising questions about peer-review calibration and AI-disclosure policies.
We limit this claim to the four ML/NLP venues covered by \scizoom and treat extension to other disciplines (mathematics, physics, biomedicine, social sciences) as an empirical question for future work, in line with the venue-stratified prevalence estimates of~\citet{liang2025quantifying}.

\paragraph{Benchmark Difficulty.}
Even DeepSeek-R1 (671B) does not outperform the best 7-8B models, and the persistent gap between lexical overlap (R-L $\sim$23-29\%) and semantic similarity (BERTScore $\sim$87-89\%) suggests that models capture core meaning but struggle to reproduce precise scientific phrasing, positioning \scizoom as a benchmark unlikely to be saturated by simple scaling.

\paragraph{Hierarchical Consistency.}
Cross-level entailment exceeds 89\% with under 0.4\% contradiction across all models, suggesting that hierarchical summarization is a viable paradigm where users can trust that TL;DRs will not introduce claims contradicted by the full abstract.

\paragraph{Applications and Future Work.}
Beyond evaluation, \scizoom can support automated paper screening at multiple granularity levels, meta-science studies tracking writing trends, and training data for domain-specific summarization models.
Future directions include fine-tuning for multi-granularity summarization, cross-domain analysis of linguistic shifts beyond ML/NLP venues, and developing era-aware models that account for Pre/Post-LLM distribution shifts.

\section{Conclusion}
\label{sec:conclusion}

We introduced \scizoom, a large-scale hierarchical benchmark comprising 44,946 papers from four premier AI conferences, with three target granularities achieving compression ratios up to 600:1.
Benchmarking seven LLMs reveals that \scizoom remains challenging even for frontier models, and cross-era evaluation underscores the importance of temporal stratification in benchmark design.
The explicit Pre/Post-LLM stratification not only enables era-aware model evaluation but also supports corpus-level analyses of scientific writing evolution, as demonstrated by our linguistic findings.
We release \scizoom publicly to facilitate future shared tasks and community benchmarking efforts.

\section*{Limitations}

While \scizoom provides the first temporally stratified, multi-granularity scientific summarization benchmark, several limitations remain.
Venue coverage differs due to OpenReview adoption timing (EMNLP: 2023 only, ICML: 2023 onward), and TL;DR availability is restricted to 47.4\%.
Our contribution extraction relies on LLM generation for 66\% of papers; although human evaluation confirms 0\% hallucination and 91.7\% full support, the generation model (Qwen3-4B) and top-performing baseline (Qwen2-7B) both belong to the Qwen family, so we release the author-written subset for circularity-sensitive evaluation.

On the evaluation side, baselines are limited to zero-shot inference with open-weight models for reproducibility; fine-tuning, few-shot prompting, and proprietary API models are left as future work.
Our human evaluation covers 50 papers (223 items), which would benefit from larger-scale annotation in future iterations.
Automatic metrics may not fully capture factual accuracy; the cross-level entailment results in Table~\ref{tab:entailment_main} take a first step toward NLI-based faithfulness scoring.

Our temporal stratification uses ChatGPT's release as a binary threshold, though LLM adoption likely followed a gradual trajectory.
All linguistic conclusions are scoped to ML/NLP; per-venue statistical tests control for venue composition, but regression with additional covariates (e.g., topic distribution, paper length) remains future work.

\bibliography{ref}

\clearpage
\appendix
\renewcommand{\thefigure}{A\arabic{figure}}
\renewcommand{\thetable}{A\arabic{table}}
\setcounter{figure}{0}
\setcounter{table}{0}

\section{Dataset Statistics and Visualization}
\label{app:stats}

\subsection{Temporal Distribution}
Tables~\ref{tab:paper_count_detail} and~\ref{tab:era} detail the venue-year distribution and era-wise split.

\begin{table}[H]
\centering
\caption{Number of papers by venue and year.}
\label{tab:paper_count_detail}
\resizebox{\linewidth}{!}{%
\begin{tabular}{lrrrrrr|r}
\toprule
Venue & 2020 & 2021 & 2022 & 2023 & 2024 & 2025 & Total \\
\midrule
NeurIPS & -- & 2,751 & 2,810 & 3,214 & 4,027 & 5,286 & 18,088 \\
ICLR & 2,212 & 2,586 & 2,612 & 3,783 & 2,258 & 3,703 & 17,154 \\
ICML & -- & -- & -- & 1,828 & 2,610 & 3,257 & 7,695 \\
EMNLP & -- & -- & -- & 2,009 & -- & -- & 2,009 \\
\midrule
\textbf{Total} & 2,212 & 5,337 & 5,422 & 10,834 & 8,895 & 12,246 & \textbf{44,946} \\
\bottomrule
\end{tabular}
}%
\end{table}

\begin{table}[H]
\centering
\caption{Dataset split by research era.}
\label{tab:era}
\resizebox{0.8\linewidth}{!}{%
\begin{tabular}{llrr}
\toprule
Era & Venue (Year) & Papers & \% \\
\midrule
\multirow{2}{*}{\prellm} & ICLR 2020--2023 & 11,193 & 24.9 \\
 & NeurIPS 2021--2022 & 5,561 & 12.4 \\
\cmidrule{2-4}
\multirow{4}{*}{\postllm} & NeurIPS 2023--2025 & 12,527 & 27.9 \\
 & ICLR 2024--2025 & 5,961 & 13.3 \\
 & ICML 2023--2025 & 7,695 & 17.1 \\
 & EMNLP 2023 & 2,009 & 4.5 \\
\midrule
\multicolumn{2}{l}{\textbf{Total}} & \textbf{44,946} & 100.0 \\
\bottomrule
\end{tabular}
}%
\end{table}

\subsection{Metadata and Topic Analysis}
Table~\ref{tab:metadata} summarizes metadata coverage, Table~\ref{tab:keywords} lists frequent keywords, and Figure~\ref{fig:dataset_structure} illustrates the hierarchical structure.

\begin{table}[H]
\centering
\caption{Metadata availability in \scizoom.}
\label{tab:metadata}
\begin{tabular}{lrr}
\toprule
Field & Count & \% \\
\midrule
Full text & 44,946 & 100.0 \\
Abstract & 44,946 & 100.0 \\
Contributions & 44,946 & 100.0 \\
Keywords & 38,878 & 86.5 \\
TL;DR & 21,295 & 47.4 \\
Presentation type & 31,949 & 71.1 \\
Primary area & 22,303 & 49.6 \\
\bottomrule
\end{tabular}
\end{table}

\begin{table}[H]
\centering
\caption{Top 10 most frequent author-provided keywords.}
\label{tab:keywords}
\begin{tabular}{clr}
\toprule
Rank & Keyword & Count \\
\midrule
1 & reinforcement learning & 2,229 \\
2 & deep learning & 1,549 \\
3 & large language models & 1,444 \\
4 & representation learning & 1,006 \\
5 & graph neural networks & 843 \\
6 & diffusion models & 792 \\
7 & large language model & 689 \\
8 & interpretability & 655 \\
9 & self-supervised learning & 634 \\
10 & federated learning & 615 \\
\bottomrule
\end{tabular}
\end{table}

\begin{figure}[H]
\centering
\scriptsize
\setlength{\tabcolsep}{3pt}
\renewcommand{\arraystretch}{1.3}

\begin{tabular}{|p{0.92\linewidth}|}
\hline
\cellcolor{black}\textcolor{white}{$G_1$: Full Paper (8,190 words)} \\
\hline
Exploring Vision Semantic Prompt for Efﬁcient Point Cloud Understanding Yixin Zha 1 Chuxin Wang 1 Wenfei Yang 1 2 Tianzhu Zhang 1 2 Feng Wu 1 2 Introduction With the growing of training data and... \textit{[8,135 words omitted]} ...8. Depth maps of ScanObjectNN-objonly. Figure 9. Depth maps of ScanObjectNN-objbg. 16 666 Figure 10. Depth maps of ScanObjectNN-hardest. 17
\\
\hline
\end{tabular}

\vspace{1mm}
\centering{$\Downarrow$}
\vspace{1mm}

\begin{tabular}{|p{0.92\linewidth}|}
\hline
\cellcolor{black}\textcolor{white}{$G_2$: Abstract (168 words)} \\
\hline
A series of pre-trained models have demonstrated promising results in point cloud understanding tasks and are widely applied to downstream tasks through fine-tuning. However, full fine-tuning leads to the forgetting of pretrained knowledge and substantial storage costs on edge devices. To address these issues, Parameter-Efficient Transfer Learning ...
\\
\hline
\end{tabular}

\begin{tabular}{|p{0.92\linewidth}|}
\hline
\cellcolor{black}\textcolor{white}{$G_3$: Contributions (82 words)} \\
\hline
(1) We propose a new paradigm that, for the ﬁrst time, leverages 2D semantic cues to improve the generalization of pretrained 3D models with minimal train... \newline
(2) We utilize 2D class tokens at multiple scales to generate prompts, and we design a Hybrid Attention Adapter to adopt efﬁcient modality fusion while ke... \newline
(3) Extensive experiments on datasets such as ScanObjectNN, ModelNet40, and ShapeNetPart demonstrate the effectiveness of our proposed paradigm.
\\
\hline
\end{tabular}

\begin{tabular}{|p{0.92\linewidth}|}
\hline
\cellcolor{black}\textcolor{white}{$G_4$: TL;DR (27 words)} \\
\hline
We propose a new paradigm that,  for the first time, leverages 2D semantics as prompts to improve the generalization of pretrained 3D models with minimal trainable parameters.
\\
\hline
\end{tabular}

\caption{Hierarchical summarization structure in \scizoom. A full paper is summarized at three granularity levels: abstract, contributions, and TL;DR.}
\label{fig:dataset_structure}
\end{figure}

\section{Contribution Extraction Pipeline}
\label{app:pipeline}

Algorithm~\ref{algo:contrib_pipeline} formalizes the pipeline described in \S\ref{sec:benchmark}; Figure~\ref{fig:contrib_prompts} shows the prompt templates.

\begin{algorithm}[t]
\small
\SetAlgoLined
\DontPrintSemicolon
\SetKwInOut{Input}{Input}
\SetKwInOut{Output}{Output}
\Input{Set of Papers $\mathcal{P} = \{P^{(1)}, ..., P^{(N)}\}$ with full text and abstract}
\Output{Set of Contributions $\mathcal{C}$}
\BlankLine
Initialize $\mathcal{C} \leftarrow \emptyset$ \;
\ForEach{$P \in \mathcal{P}$}{
    \tcp{Stage 1: Rule-based Extraction}
    $S \leftarrow \text{FindContribSection}(P)$ \tcp*[r]{Regex patterns}
    $C_{rule} \leftarrow \text{ExtractBulletPoints}(S)$ \;
    \BlankLine
    \If{$|C_{rule}| \geq 2$}{
        \tcp{Stage 2: LLM Validation}
        $ctx \leftarrow \text{GetContextBefore}(S)$ \;
        $valid \leftarrow \text{ValidateLLM}(ctx, C_{rule})$ \tcp*[r]{Qwen3-4B}
        \If{$valid$}{
            $\mathcal{C}.\text{add}(C_{rule}, \text{source}=\text{`rule\_validated'})$ \;
            \textbf{continue} \;
        }
    }
    \BlankLine
    \tcp{Stage 3: LLM Generation (Fallback)}
    $intro \leftarrow \text{ExtractIntroLastPara}(P)$ \;
    $C_{gen} \leftarrow \text{GenerateLLM}(P.abstract, intro)$ \tcp*[r]{Qwen3-4B}
    $\mathcal{C}.\text{add}(C_{gen}, \text{source}=\text{`llm\_generated'})$ \;
}
\Return $\mathcal{C}$ \;
\caption{3-Stage Contribution Extraction Pipeline}
\label{algo:contrib_pipeline}
\end{algorithm}

\begin{figure}[H]
    \centering
    \begin{tcolorbox}[
        colback=white, colframe=black, boxrule=0.5pt, arc=2pt,
        left=3pt, right=3pt, top=2pt, bottom=2pt,
        title={\small\textbf{\sffamily Stage 2: Validation Prompt}},
        colbacktitle=black, coltitle=white, fonttitle=\sffamily,
        toptitle=1pt, bottomtitle=1pt
    ]
    \scriptsize \ttfamily
    Context: "\{context\_before\}"\\[2pt]
    Extracted items: - \{contribution 1\} - \{contribution 2\} ...\\[2pt]
    Are these valid research contributions? Answer only "yes" or "no".
    \end{tcolorbox}
    \vspace{-2pt}
    \begin{tcolorbox}[
        colback=white, colframe=black, boxrule=0.5pt, arc=2pt,
        left=3pt, right=3pt, top=2pt, bottom=2pt,
        title={\small\textbf{\sffamily Stage 3: Generation Prompt}},
        colbacktitle=black, coltitle=white, fonttitle=\sffamily,
        toptitle=1pt, bottomtitle=1pt
    ]
    \scriptsize \ttfamily
    Abstract: \{abstract\}\\[2pt]
    Introduction: \{intro\_last\_para\}\\[2pt]
    Extract 2-5 key research contributions. Output as JSON array: {[}"contribution 1", "contribution 2", ...{]}
    \end{tcolorbox}
    \caption{Prompt templates for contribution extraction with Qwen3-4B-Instruct.}
    \label{fig:contrib_prompts}
\end{figure}

\section{Input Preprocessing and Inference Details}
\label{app:preproc}

For inputs exceeding a model's context window, we apply \emph{front-back truncation}: we keep the leading $L_h$ tokens and trailing $L_t$ tokens ($L_h{=}L_t$) and drop the middle, preserving the introduction and conclusion.
We use greedy decoding with generation budgets of 512 tokens (Abstract/Contributions) and 64 tokens (TL;DR); all templates and parameters are included with the release.

\section{Cross-Era Evaluation Details}
\label{app:cross_era}

Table~\ref{tab:cross_era_full} presents complete cross-era results with all six metrics.

\begin{table*}[t]
\centering
\caption{Cross-era evaluation on \scizoom. Performance comparison on \prellm and \postllm test sets. \textbf{Bold}: best, \underline{underline}: second-best per column.}
\label{tab:cross_era_full}
\resizebox{\textwidth}{!}{%
\begin{tabular}{lcl|cccccc|cccccc}
\toprule
& & & \multicolumn{6}{c|}{\textbf{\prellm}} & \multicolumn{6}{c}{\textbf{\postllm}} \\
\textbf{Task} & & \textbf{Model} & \textbf{R-1} & \textbf{R-2} & \textbf{R-L} & \textbf{BL4} & \textbf{MTR} & \textbf{BS} & \textbf{R-1} & \textbf{R-2} & \textbf{R-L} & \textbf{BL4} & \textbf{MTR} & \textbf{BS} \\
\midrule
\multirow{6}{*}{Abstract}
& \multirow{3}{*}{\rotatebox{90}{\postllm}}
& Mistral-7B-v0.3  & \textbf{43.1} & \textbf{16.1} & \textbf{23.9} & \textbf{4.9} & \textbf{21.4} & 86.7          & 42.3 & 15.4 & 23.3 & \textbf{4.4} & \textbf{20.5} & 86.7 \\
& & Llama-3.1-8B     & 40.6 & \underline{14.9} & 22.7 & 3.5 & 18.9 & 86.8                                          & 40.3 & 14.6 & 22.2 & 3.4 & 18.6 & 86.8 \\
& & Qwen2-7B         & 41.8 & 14.8 & 23.0 & 3.6 & 19.6 & \underline{86.9}                                           & \textbf{42.6} & \underline{15.6} & \underline{23.5} & 3.8 & 20.0 & \underline{87.1} \\
\cmidrule{3-15}
& \multirow{3}{*}{\rotatebox{90}{\prellm}}
& Mistral-7B-v0.1  & 33.6 & 12.8 & 19.5 & 3.3 & 16.3 & 85.5                                                        & 30.0 & 11.2 & 17.4 & 2.5 & 13.8 & 85.1 \\
& & Llama-3-8B       & \underline{42.4} & \textbf{16.1} & \underline{23.7} & \underline{4.1} & \underline{20.3} & \textbf{87.1}  & \underline{42.5} & \textbf{16.5} & \textbf{23.9} & \underline{4.1} & \underline{20.3} & \textbf{87.2} \\
& & Qwen-7B          & 36.4 & 12.0 & 20.0 & 3.3 & 17.5 & 85.5                                                        & 32.3 & 10.5 & 17.7 & 2.6 & 15.1 & 85.1 \\
\midrule
\multirow{6}{*}{Contrib}
& \multirow{3}{*}{\rotatebox{90}{\postllm}}
& Mistral-7B-v0.3  & 33.7 & 14.5 & 23.6 & 3.4 & 16.5 & 86.2                                                        & 33.5 & 14.9 & 23.7 & 3.6 & 16.5 & 86.2 \\
& & Llama-3.1-8B     & \underline{37.8} & \underline{16.2} & \underline{25.7} & 4.4 & 18.7 & \underline{86.9}        & \underline{37.8} & 16.3 & \underline{25.7} & \underline{4.7} & \underline{18.8} & \underline{86.9} \\
& & Qwen2-7B         & \textbf{40.9} & \textbf{17.9} & \textbf{27.9} & \textbf{5.3} & \textbf{21.0} & \textbf{87.3} & \textbf{43.2} & \textbf{21.0} & \textbf{30.9} & \textbf{7.4} & \textbf{23.6} & \textbf{87.8} \\
\cmidrule{3-15}
& \multirow{3}{*}{\rotatebox{90}{\prellm}}
& Mistral-7B-v0.1  & 36.8 & 15.0 & 24.0 & \underline{4.8} & \underline{19.0} & 86.2  & 36.1 & 14.6 & 23.6 & \underline{4.7} & 18.6 & 86.1 \\
& & Llama-3-8B       & 32.0 & 14.6 & 23.1 & 3.0 & 15.2 & 86.2                                                        & 33.6 & \underline{16.8} & 25.2 & 4.1 & 16.8 & 86.4 \\
& & Qwen-7B          & 22.8 &  8.3 & 15.6 & 1.1 &  9.9 & 84.3                                                        & 22.4 &  8.1 & 15.2 & 1.1 &  9.6 & 84.2 \\
\midrule
\multirow{6}{*}{TL;DR}
& \multirow{3}{*}{\rotatebox{90}{\postllm}}
& Mistral-7B-v0.3  & 36.1 & \underline{15.6} & 27.8 & \underline{4.8} & 27.4 & 88.9  & 36.6 & 16.0 & 28.2 & 5.0 & 27.2 & \underline{89.0} \\
& & Llama-3.1-8B     & \underline{36.5} & \textbf{16.4} & \underline{28.3} & \textbf{5.1} & \underline{27.8} & \textbf{89.1}  & \underline{37.3} & \underline{16.6} & \underline{28.8} & \underline{5.2} & \underline{27.7} & \textbf{89.1} \\
& & Qwen2-7B         & 34.3 & 14.3 & 25.6 & 3.7 & 22.8 & 88.3                                                        & 35.4 & 15.1 & 26.7 & 4.1 & 23.3 & 88.5 \\
\cmidrule{3-15}
& \multirow{3}{*}{\rotatebox{90}{\prellm}}
& Mistral-7B-v0.1  & 32.7 & 14.2 & 24.8 & 4.4 & \textbf{29.2} & 88.2                                                & 33.5 & 14.4 & 25.4 & 4.5 & \textbf{28.6} & 88.3 \\
& & Llama-3-8B       & \textbf{37.1} & \textbf{16.4} & \textbf{28.4} & \textbf{5.1} & 27.4 & \underline{89.0}        & \textbf{37.7} & \textbf{16.8} & \textbf{29.0} & \textbf{5.3} & 27.6 & \underline{89.0} \\
& & Qwen-7B          & 32.3 & 13.7 & 24.4 & 4.0 & 25.6 & 88.0                                                        & 33.4 & 14.6 & 25.7 & 4.5 & 25.8 & 88.3 \\
\bottomrule
\end{tabular}}
\end{table*}

\section{Domain-Specific Summarization Metrics}
\label{app:domain_metrics}

Tables~\ref{tab:domain_abstract}--\ref{tab:domain_tldr} report CIDEr~\cite{vedantam2014consensus}, SciBERTScore~\cite{beltagy2019scibert}, and BARTScore~\cite{yuan2021bartscore} per task.
DeepSeek-R1's comparable SciBERTScore but lower BARTScore confirms that its reduced lexical overlap reflects verbose style rather than degraded semantic quality.

\begin{table}[H]
\centering
\caption{Abstract generation -- domain-specific metrics.}
\label{tab:domain_abstract}
\begin{tabular}{lccc}
\toprule
\textbf{Model} & \textbf{CIDEr} & \textbf{SciBERT} & \textbf{BARTScore} \\
\midrule
Llama-3.1-8B    & 0.011 & 68.0 & $-$2.95 \\
Llama-3-8B      & 0.014 & \textbf{69.0} & $-$2.91 \\
Mistral-v0.3    & \textbf{0.030} & 67.9 & $-$3.02 \\
Qwen2-7B        & 0.017 & 68.6 & $-$3.10 \\
DeepSeek-R1     & 0.016 & 68.1 & $-$3.38 \\
\bottomrule
\end{tabular}
\end{table}

\begin{table}[H]
\centering
\caption{Contribution extraction -- domain-specific metrics.}
\label{tab:domain_contrib}
\begin{tabular}{lccc}
\toprule
\textbf{Model} & \textbf{CIDEr} & \textbf{SciBERT} & \textbf{BARTScore} \\
\midrule
Llama-3.1-8B    & 0.069 & 65.7 & $-$3.30 \\
Qwen2-7B        & \textbf{0.110} & \textbf{67.7} & \textbf{$-$3.20} \\
DeepSeek-R1     & 0.034 & 67.3 & $-$3.64 \\
\bottomrule
\end{tabular}
\end{table}

\begin{table}[H]
\centering
\caption{TL;DR generation -- domain-specific metrics.}
\label{tab:domain_tldr}
\resizebox{0.85\linewidth}{!}{%
\begin{tabular}{lccc}
\toprule
\textbf{Model} & \textbf{CIDEr} & \textbf{SciBERT} & \textbf{BARTScore} \\
\midrule
Llama-3.1-8B    & \textbf{0.462} & \textbf{71.0} & \textbf{$-$3.74} \\
Qwen2-7B        & 0.398 & 69.5 & $-$4.24 \\
DeepSeek-R1     & 0.369 & 68.9 & $-$4.69 \\
\bottomrule
\end{tabular}}
\end{table}

\section{Cross-Level Entailment: Full Results}
\label{app:entailment}

Table~\ref{tab:entailment_abs2contrib} reports Abstract$\rightarrow$Contribution entailment; contradiction remains $\leq 0.4\%$ across all models.

\begin{table}[H]
\centering
\caption{Cross-level entailment for Abstract$\rightarrow$Contribution (\%).}
\label{tab:entailment_abs2contrib}
\resizebox{0.85\linewidth}{!}{%
\begin{tabular}{lccc}
\toprule
\textbf{Model} & \textbf{Entail} & \textbf{Contradict} & \textbf{Neutral} \\
\midrule
Mistral-7B-v0.3   & 86.3 & 0.1 & 13.6 \\
Llama-3-8B        & 75.1 & 0.1 & 24.8 \\
Llama-3.1-8B      & 68.6 & 0.2 & 31.2 \\
Qwen2-7B          & 59.2 & 0.2 & 40.6 \\
DeepSeek-R1       & 56.6 & 0.4 & 42.9 \\
\bottomrule
\end{tabular}}
\end{table}

\section{Per-Venue Linguistic-Shift Statistics}
\label{app:per_venue}

Table~\ref{tab:per_venue} reports per-venue formulaic-phrase ratios and hedging changes with Mann-Whitney U statistics~\cite{mann1947test}; both venues confirm the corpus-wide trend at $p{<}10^{-15}$.

\begin{table}[H]
\centering
\caption{Per-venue linguistic shifts. ``Formulaic'' = ratio of mean trigram frequency \postllm vs.\ \prellm.}
\label{tab:per_venue}
\resizebox{\linewidth}{!}{%
\begin{tabular}{lcccc}
\toprule
\textbf{Venue} & $N$ Pre & $N$ Post & \textbf{Formulaic} & \textbf{Hedging} \\
\midrule
NeurIPS   & 5,561  & 12,527 & 1.76$\times$ & $-31.1\%$ \\
ICLR      & 11,193 & 5,961  & 2.21$\times$ & $-20.7\%$ \\
\midrule
\multicolumn{5}{c}{\textbf{Mann-Whitney U statistics}} \\
\midrule
NeurIPS (formulaic) & \multicolumn{4}{c}{$U{=}3.10{\times}10^{7}$,\quad $p{<}10^{-56}$} \\
NeurIPS (hedging)   & \multicolumn{4}{c}{$U{=}3.88{\times}10^{7}$,\quad $p{<}10^{-46}$} \\
ICLR (formulaic)    & \multicolumn{4}{c}{$U{=}2.81{\times}10^{7}$,\quad $p{<}10^{-127}$} \\
ICLR (hedging)      & \multicolumn{4}{c}{$U{=}3.55{\times}10^{7}$,\quad $p{<}10^{-15}$} \\
\midrule
Corpus (formulaic)  & \multicolumn{4}{c}{$U{=}2.09{\times}10^{8}$,\quad $p{<}10^{-177}$,\quad $r{=}0.117$} \\
Corpus (hedging)    & \multicolumn{4}{c}{$U{=}2.55{\times}10^{8}$,\quad $p{<}10^{-63}$,\quad $r{=}-0.081$} \\
\bottomrule
\end{tabular}}
\end{table}

\section{Retrieval Metric Definitions and Full Results}
\label{app:retrieval_def}

For all retrieval tables, let $\mathcal{Q}$ be the set of queries and $\mathrm{rank}(q)$ the rank of the ground-truth target (1 = top). Then

\begin{equation}
\mathrm{R@}K = \frac{1}{|\mathcal{Q}|} \sum_{q \in \mathcal{Q}} \mathbb{1}\!\left[\mathrm{rank}(q) \leq K\right]
\end{equation}
\begin{equation}
\mathrm{MRR} = \frac{1}{|\mathcal{Q}|} \sum_{q \in \mathcal{Q}} \frac{1}{\mathrm{rank}(q)}
\end{equation}

Embeddings are produced by NV-Embed-v2~\cite{lee2025nv}; we rank by cosine similarity.
Table~\ref{tab:retrieval_gen} reports model-generated retrieval results.

\begin{table*}[t]
\centering
\caption{Cross-granularity retrieval on model-generated summaries (\%). \textbf{Bold}: best, \underline{underline}: second-best.}
\label{tab:retrieval_gen}
\resizebox{\textwidth}{!}{%
\begin{tabular}{ll|cccc|cccc|cccc}
\toprule
& & \multicolumn{4}{c|}{\textbf{Mistral-7B}} & \multicolumn{4}{c|}{\textbf{Llama-3-8B}} & \multicolumn{4}{c}{\textbf{Qwen2-7B}} \\
\textbf{Group} & \textbf{Direction} & R@1 & R@5 & R@10 & MRR & R@1 & R@5 & R@10 & MRR & R@1 & R@5 & R@10 & MRR \\
\midrule
\multirow{6}{*}{\rotatebox{90}{\textbf{Gen Internal}}}
& tldr$\rightarrow$abs    & \underline{98.4} & 99.6 & 99.8 & 99.0 & 98.1 & 99.8 & 99.9 & 98.9 & \textbf{99.3} & 99.9 & 99.9 & 99.6 \\
& tldr$\rightarrow$cont   & \underline{87.2} & 93.1 & 94.6 & 89.9 & 81.8 & 90.4 & 92.6 & 85.7 & \textbf{92.8} & 96.7 & 97.7 & 94.6 \\
& cont$\rightarrow$abs    & \underline{95.3} & 98.3 & 98.7 & 96.6 & 94.6 & 97.9 & 98.4 & 96.1 & \textbf{98.2} & 99.5 & 99.6 & 98.8 \\
& cont$\rightarrow$tldr   & \underline{87.0} & 93.1 & 94.4 & 89.8 & 73.3 & 84.0 & 86.8 & 78.2 & \textbf{90.3} & 95.4 & 96.8 & 92.7 \\
& abs$\rightarrow$tldr    & 97.6 & 99.5 & 99.8 & 98.5 & \underline{97.8} & 99.7 & 99.8 & 98.6 & \textbf{98.3} & 99.7 & 99.9 & 99.0 \\
& abs$\rightarrow$cont    & \underline{92.0} & 96.2 & 97.3 & 94.0 & 90.9 & 96.1 & 97.2 & 93.3 & \textbf{96.1} & 98.6 & 99.1 & 97.3 \\
\midrule
\multirow{5}{*}{\rotatebox{90}{\textbf{Gen$\rightarrow$GT}}}
& gen\_abs$\rightarrow$gt\_abs         & 96.9 & 99.7 & 99.8 & 98.2 & \underline{98.0} & 99.8 & 99.9 & 98.8 & \textbf{98.7} & 99.9 & 99.9 & 99.2 \\
& gen\_tldr$\rightarrow$gt\_abs        & 84.1 & 94.3 & 96.4 & 88.5 & \underline{86.5} & 95.6 & 97.4 & 90.5 & \textbf{92.7} & 98.0 & 98.8 & 95.1 \\
& gen\_cont$\rightarrow$gt\_abs        & \underline{77.9} & 89.3 & 91.7 & 83.1 & 75.3 & 88.6 & 91.8 & 81.3 & \textbf{91.8} & 97.2 & 98.3 & 94.2 \\
& gen\_tldr$\rightarrow$gt\_tldr       & 66.9 & 80.7 & 84.6 & 73.3 & \underline{71.4} & 84.1 & 87.8 & 77.2 & \textbf{74.6} & 86.4 & 89.4 & 79.9 \\
& gen\_cont$\rightarrow$gt\_cont       & \underline{76.5} & 88.2 & 91.0 & 81.8 & 73.3 & 87.7 & 91.4 & 79.8 & \textbf{91.4} & 96.3 & 97.5 & 93.7 \\
\bottomrule
\end{tabular}}
\end{table*}

\begin{figure*}[t]
\centering
\scriptsize
\setlength{\tabcolsep}{3pt}
\renewcommand{\arraystretch}{1.3}

\begin{tabular}{|p{0.30\textwidth}|p{0.42\textwidth}|p{0.20\textwidth}|}
\hline
\multicolumn{3}{|l|}{\cellcolor{black}\textcolor{white}{Full Paper (8,190 words)}} \\
\hline
\multicolumn{3}{|p{0.96\textwidth}|}{
Exploring Vision Semantic Prompt for Efﬁcient Point Cloud Understanding Yixin Zha 1 Chuxin Wang 1 Wenfei Yang 1 2 Tianzhu Zhang 1 2 Feng Wu 1 2 Introduction With the growing of training data and model parameters, large foundation models... \textit{[8,125 words omitted]} ...of Modelnet40. 15 666 Figure 8. Depth maps of ScanObjectNN-objonly. Figure 9. Depth maps of ScanObjectNN-objbg. 16 666 Figure 10. Depth maps of ScanObjectNN-hardest. 17
} \\
\hline
\hline
\centering\textbf{Abstract} & \centering\textbf{Contributions} & \centering\arraybackslash\textbf{TL;DR} \\
\hline
\hline
\multicolumn{3}{|l|}{\cellcolor{black}\textcolor{white}{Ground Truth}} \\
\hline
A series of pre-trained models have demonstrated promising results in point cloud understanding tasks and are widely applied to downstream tasks through fine-tuning. However, full fine-tuning leads to the forgetting of pretrained knowledge and substa...
&
(1) We propose a new paradigm that, for the ﬁrst time, leverages 2D semantic cues to improve t... \newline
(2) We utilize 2D class tokens at multiple scales to generate prompts, and we design a Hybrid ... \newline
(3) Extensive experiments on datasets such as ScanObjectNN, ModelNet40, and ShapeNetPart demon...
&
We propose a new paradigm that,  for the first time, leverages 2D semantics as prompts to improve the generalization of pretrained 3D models with minimal trainable parameters.
\\
\hline
\hline
\multicolumn{3}{|l|}{\cellcolor{black}\textcolor{white}{Mistral-7B}} \\
\hline
The paper explores the use of vision semantic prompts for efficient point cloud understanding. It addresses the limitations of full fine-tuning strategies in point cloud understanding tasks by proposing a series of Param...
&
(1) Proposes a method to address limitations of full fine-tuning strategies in point... \newline
(2) Investigates the effectiveness of incorporating self-attention mechanisms within... \newline
(3) Investigates the impact of 2D semantic cues...
&
The paper proposes a method to integrate 2D semantic cues with 3D features for efficient point cloud understanding, and investigates the effectiveness of self-attention mechanisms and 2D semantic cues.
\\
\hline
\hline
\multicolumn{3}{|l|}{\cellcolor{black}\textcolor{white}{Llama-3-8B}} \\
\hline
This paper proposes a novel paradigm that leverages visual semantic prompts to improve the generalization of pretrained 3D models with minimal trainable parameters. The approach utilizes 2D depth maps from three orthogon...
&
(1) Proposing a novel paradigm for leveraging visual semantic prompts to improve 3D ... \newline
(2) Introducing a multi-scale semantic cues injection strategy for modality fusion... \newline
(3) Designing a hybrid attention adapter for efficient modality fusion...
&
The paper proposes a novel approach that uses 2D visual cues to improve the generalization of 3D models with minimal trainable parameters.
\\
\hline
\hline
\multicolumn{3}{|l|}{\cellcolor{black}\textcolor{white}{Qwen2-7B}} \\
\hline
This paper introduces a novel paradigm for enhancing the generalization of pretrained 3D models through the integration of 2D semantic cues, while maintaining parameter efficiency. The proposed method, leveraging 3D-to-2...
&
(1) Proposes a new paradigm that leverages 2D semantic cues to improve the generaliz... \newline
(2) Utilizes 2D class tokens at multiple scales to generate prompts and designs a Hy... \newline
(3) Demonstrates the effectiveness of the proposed paradigm through extensive experi...
&
A novel 3D model enhancement paradigm using 2D semantic cues for improved generalization and parameter efficiency.
\\
\hline
\end{tabular}

\caption{Qualitative comparison on ICML 2025 paper ``Exploring Vision Semantic Prompt for Efficient Poi...''. Ground truth via rule-based extraction.}
\label{fig:qualitative_comparison}
\end{figure*}

\end{document}